\documentclass[10pt,twocolumn,letterpaper]{article}

\usepackage{cvpr}              

\usepackage[dvipsnames]{xcolor}

\definecolor{cvprblue}{rgb}{0.21,0.49,0.74}
\usepackage[pagebackref,breaklinks,colorlinks,citecolor=cvprblue]{hyperref}

\usepackage{graphicx}
\usepackage{amsmath}
\usepackage{amssymb}
\usepackage{booktabs}
\usepackage{multirow}
\usepackage{color}
\usepackage{colortbl}
\usepackage{bbding}
\usepackage{amsmath}
\usepackage[accsupp]{axessibility}

\def\paperID{3066} 
\def\confName{CVPR}
\def\confYear{2025}

\title{Learning to Detect Objects from 
Multi-Agent \\ LiDAR Scans without Manual Labels
}

\author{Qiming Xia\textsuperscript{\rm 1}\textsuperscript{\rm 2} \quad Wenkai Lin\textsuperscript{\rm 1}\textsuperscript{\rm 2} \quad Haoen Xiang\textsuperscript{\rm 1}\textsuperscript{\rm 2} \quad Xun Huang\textsuperscript{\rm 1}\textsuperscript{\rm 2} \quad \\ Siheng Chen\textsuperscript{\rm 3} \quad Zhen Dong\textsuperscript{\rm 4} \quad Cheng Wang\textsuperscript{\rm 1}\textsuperscript{\rm 2} \quad Chenglu Wen\textsuperscript{\rm 1}\textsuperscript{\rm 2}\thanks{Corresponding author, {\tt\small clwen@xmu.edu.cn}} \\
    \textsuperscript{\rm 1}Fujian Key Laboratory of Sensing and Computing for Smart Cities, Xiamen University, China \\
    \textsuperscript{\rm 2}Key Laboratory of Multimedia Trusted Perception and Efficient Computing, \\Ministry of Education of China, Xiamen University, China \\
    \textsuperscript{\rm 3}Shanghai Jiao Tong University 
    \textsuperscript{\rm 4}Wuhan University 
\\
}

\begin{document}
\maketitle

\begin{abstract}
Unsupervised 3D object detection serves as an important solution for offline 3D object annotation. However, due to the data sparsity and limited views, the clustering-based label fitting in unsupervised object detection often generates low-quality pseudo-labels.
Multi-agent collaborative dataset, which involves the sharing of complementary observations among agents, holds the potential to break through this bottleneck. In this paper, we introduce a novel unsupervised method that learns to \textbf{D}etect \textbf{O}bjects from Mul\textbf{t}i-\textbf{A}gent LiDAR scans, termed~\texttt{DOtA}, without using labels from external. \texttt{DOtA} first uses the internally shared ego-pose and ego-shape of collaborative agents to initialize the detector, leveraging the generalization performance of neural networks to infer preliminary labels. Subsequently,
\texttt{DOtA} uses the complementary observations between agents to perform multi-scale encoding on preliminary labels, then decodes high-quality and low-quality labels.  
These labels are further used as prompts to guide a correct feature learning process, thereby enhancing the performance of the unsupervised object detection task. 
Extensive experiments on the V2V4Real and OPV2V datasets show that our~\texttt{DOtA} outperforms state-of-the-art unsupervised 3D object detection methods. Additionally, we also validate the effectiveness of the~\texttt{DOtA} labels under various collaborative perception frameworks.
The code is available at \url{https://github.com/xmuqimingxia/DOtA}.

\end{abstract}
\section{Introduction}

\begin{figure}[t]
  \centering
   \includegraphics[width=1\linewidth]{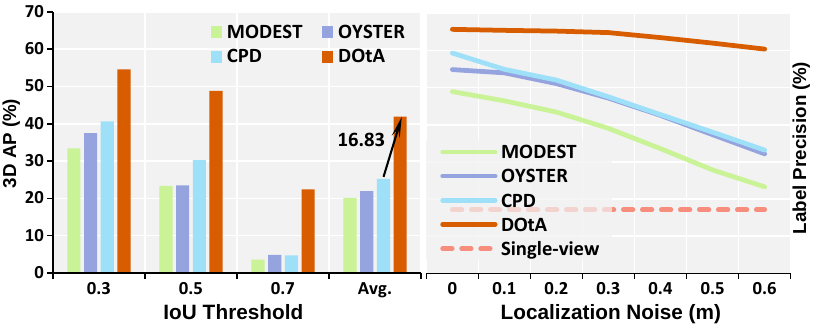}
   \caption{Comparison of the performance of various methods under multi-view synchronous observation. (a) On the left, Our \texttt{DOtA} achieves best performance on the real-world collaborative V2V4Real~\cite{v2v4real} dataset (more details are in Tab.~\ref{tab:com_unsupervised}). (b) On the right, we adhere to the localization noise parameters established in Where2comm\cite{where2comm}, and conducted experiments on the simulation dataset OPV2V~\cite{opv2v} to assess the robustness of our approach against realistic localization noise. \texttt{DOtA} is more robust to the localization noise than previous SOTAs.
   }
   \label{fig:spaesely-supervised}
\end{figure}

\begin{figure}[t]
  \centering
   \includegraphics[width=1\linewidth]{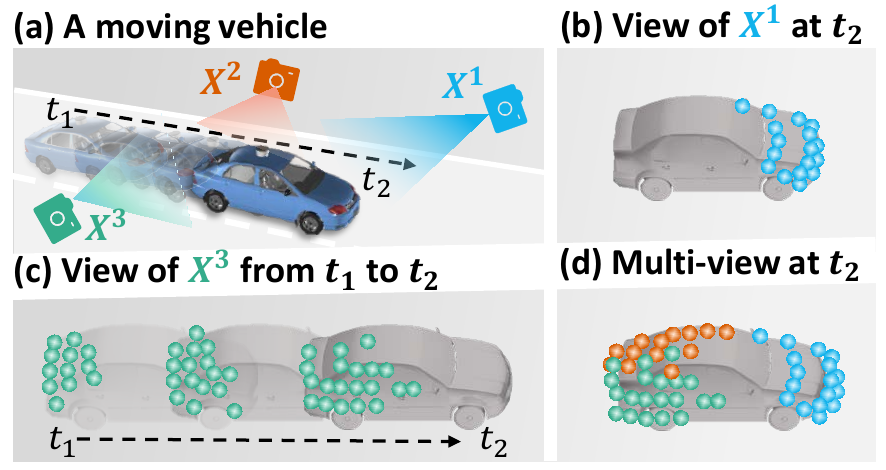}
   \caption{(a) A toy example demonstrating a moving vehicle under multi-agent observation. (b) The instance structure is incomplete under single-view observation in the current frame; (c) Historical frame from a single-view cannot complete missing information; (d) Multi-view observation for a moving vehicle.}
   \label{fig:experiment-setting}
\end{figure}

Learning-based 3D object perception, a key autonomous driving task, has recently seen rapid development in both the industrial and academic fields. A critical component of it is the \textit{offline} 3D object automatic annotation, which generates accurate labels for unlabeled data~\cite{ODNL}. 
Depending on the specific requirements, there are many different offline schemes.
Among them, one extreme emphasizes raising the upper bound of label quality, which uses a pre-trained detector and information from future frames \cite{detzero, offboard, auto4d, ODNL} to infer accurate labels. However, this strategy relies on costly manual labeling to pre-train a powerful detector~\cite{mvf, virtual, pvrcnn++, bevnext}, which is too expensive to scale up. 
Other solutions, abandoning manual labeling, leverage common-sense information to generate labels from the point cloud distribution~\cite{cpd, rffm, MODEST, OYSTER}.
However, due to the data sparsity and limited views of LiDAR scans, the distribution of foreground points is often incomplete, especially for moving objects (Fig.~\ref{fig:experiment-setting} (b,c)), severely impacting the performance of these unsupervised methods.

 
In this paper, we propose a novel orthogonal approach of improving unsupervised 3D object detection by exploring the potential of multi-agent collaboration. We assume that multi-agents are conducting synchronized observations from multi-view within the same scene, with each agent sharing their pose and shape with the others. This would bring two outstanding benefits: \textit{i)} synchronized observations from multi-view of multi-agent are adept at significantly  completing the missing point cloud distribution (Fig.~\ref{fig:experiment-setting} (d)); \textit{ii)} 
from the \textit{unlabeled} shared information (ego-pose \& ego-shape), we can obtain bounding box descriptions of the agents at no cost, and this subtle signal can support training a weak initial detector and generate preliminary labels. This compensates for the shortcomings of traditional unsupervised label generation methods that rely heavily on the complete distribution of foreground points~\cite{OYSTER}.

However, 
\textit{unlabeled} synchronized observation data also brings two new challenges to the unsupervised 3D object detection task: \textit{i)}
the initial detector trained with shared information cannot ensure label quality due to a large number of false positives, as shown in Tab.~\ref{tab:motivation_1};
\textit{ii)} 
communication delays and localization noise of agents disrupt the alignment of synchronized observation~\cite{coca, v2v4real}, 
significantly degrading the performance of traditional unsupervised object detection algorithms (Fig.~\ref{fig:spaesely-supervised}).
Therefore, agents need a more robust way to utilize synchronized observation to generate high-quality labels.

Following this design rationale, we propose an unsupervised 3D detection method that learns to Detect Objects from Multi-Agent LiDAR scans, termed~\texttt{DOtA}. It includes three parts: \textit{i)} preliminary label generation, which pre-trained the initial detector with shared information to infer labels; \textit{ii)} preliminary label refinement, which emphasizes the consistency of local point cloud multi-scale distribution across multi-view to eliminate false positive labels; and \textit{iii)} label-internal contrastive learning, which uses refined labels as cues to encourage correct feature learning and suppress erroneous feature learning. Compared with traditional unsupervised 3D object detection methods\cite{OYSTER,cpd},~\texttt{DOtA} designs a novel label refinement and detector training scheme.


The effectiveness of our design is verified both on a real-world dataset, V2V4Real~\cite{v2v4real}, and a simulation dataset, OPV2V~\cite{opv2v}. As shown in Fig.~\ref{fig:spaesely-supervised}, our \texttt{DOtA} outperformed the previous SOTA unsupervised methods average performances by $16.83\%$ on the real-world V2V4Real dataset; meanwhile, we conducted simulations of realistic localization noise on the OPV2V dataset, and the labels inferred by~\texttt{DOtA} demonstrated the strongest robustness. 
The main contributions of this paper are as follows:

\begin{itemize}
\item We introduce the first unsupervised 3D object detection method (\texttt{DOtA}) derived from multi-agent LiDAR scans. \texttt{DOtA} does not rely on external labels; instead, it only leverages unlabeled internally shared information among the agents.

\item We design a multi-scale bounding-box encoding module that leverages the complementary observation shared among multiple agents to discern high-quality labels from preliminary labels.
\item We propose a label-internal contrastive learning method, which takes the discernible preliminary labels as prompts, encouraging positive predictions while reducing the occurrence of false positives.
\end{itemize}

\section{Related Work}

\begin{figure*}[t]
  \centering
   \includegraphics[width=0.99\linewidth]{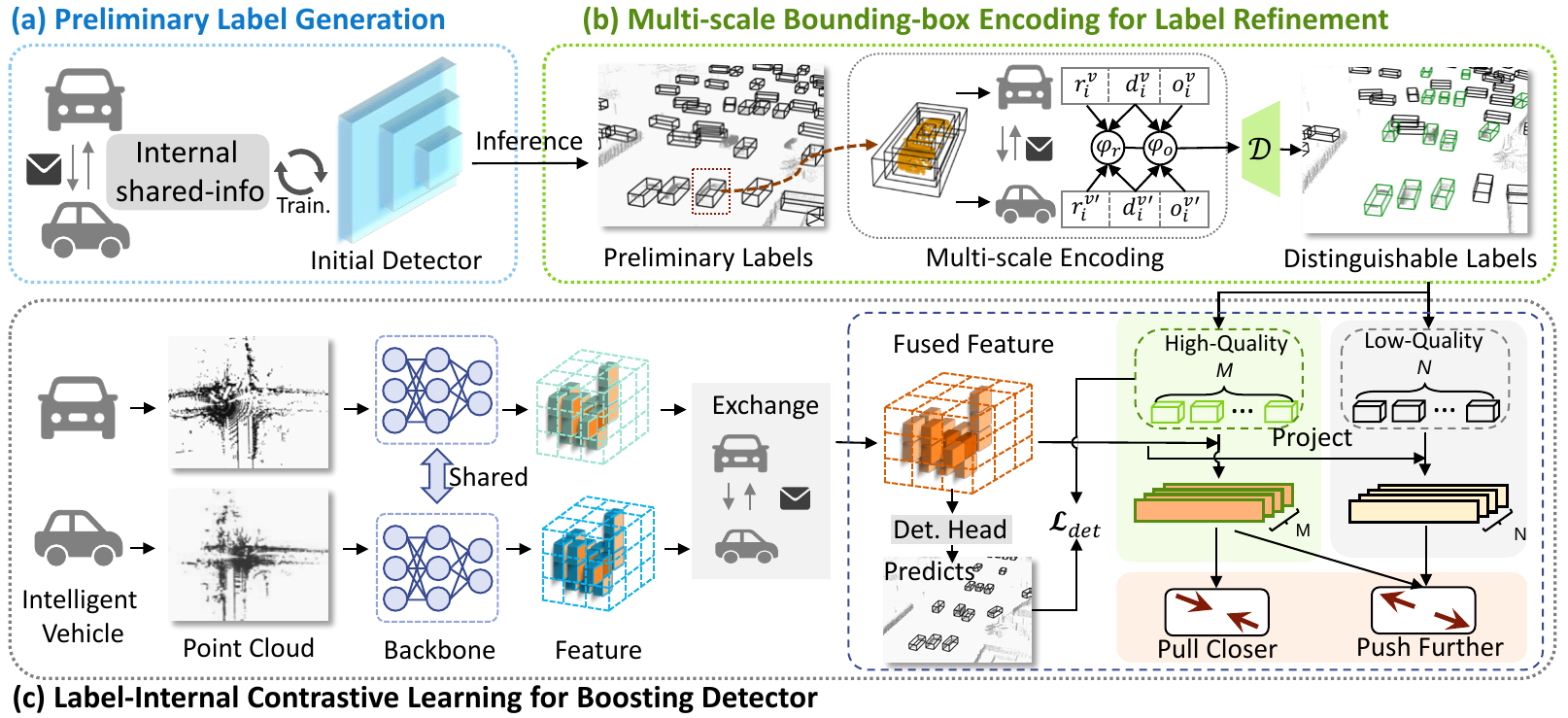}
   \caption{The overview of proposed \texttt{DOtA}. (a) The initial detector, pre-trained with shared information, infer preliminary labels. (b) Multi-scale transformations are utilized to encode contextual information for preliminary labels, with the discriminator $\mathcal{D}$ integrating the encoded information from various agents to distinguish between high-quality and low-quality labels. (c) Distinguishable labels serve as prompts, and Label-Internal Contrastive Learning (\textbf{LICL}) is leveraged to guide the learning of correct features while suppressing the learning of erroneous ones.}
   \label{fig:framework}
\end{figure*}

\subsection{Multi-agent collaborative perception}
Collaborative perception~\cite{where2comm,CoCa3D,DiscoGraph,v2x-sim,when2com,who2com,CoAlign,v2vnet,v2xp-asg,opv2v,cobevt,dair-v2x, V2X-ViTv2} is an emerging application of multi-agent communication systems in perception tasks. 
It enables different agents to share complementary perception information, expanding their view and eliminating blind spots.
Currently, collaborative perception systems have made significant progress in improving perception performance~\cite{cobevt,v2x-vit} and robustness on practical issues, such as communication bandwidth constraints~\cite{codefilling}, pose error~\cite{CoAlign,ccpe} and latency ~\cite{SyncNet,CoBEVFlow}. 
However, the cost of annotating collaborative perception datasets is extremely high. Compared to traditional single-agent perception datasets, the annotation cost of collaborative perception increases linearly with the number of agents~\cite{v2v4real}.
This significantly hinders the progress of collaborative perception algorithms on real-world data.
To address this problem, we propose an unsupervised method to avoid manual labeling by only using internally shared information in collaborative communication, thus releasing data annotation costs.

\subsection{Unsupervised 3D object detection}
In the field of single-agent 3D object detection, researchers are increasingly turning their attention to discovering 3D objects without manual annotation. 
MODEST~\cite{MODEST} first incorporates common-sense information~\cite{comm-1, comm-2, comm-3, comm-4}, using the ephemerality of repeatedly scanned point clouds to identify moving objects in the scene. 
Building on this foundation, 
many unsupervised methods~\cite{motionun, OYSTER, rffm, cpd, liso} explore additional constraints to generate higher-quality pseudo-labels. 
At the same time, some methods~\cite{4dun, union} begin to explore unsupervised 3D object detection based on multi-modal data.
However, constrained by a single viewpoint and the sparsity of point clouds, existing unsupervised object detection methods struggle to model objects, particularly those that are in motion.
To break through this bottleneck, \texttt{DOtA} attempts to conduct research on 3D object detection without manual annotation using a multi-view observational collaborative dataset.

\subsection{Label-efficient 3D object detection}
In addition, label-efficient 3D object detection methods consider the detection task from the view of weak supervision, exploring the use of low-cost manual annotation to approach the performance of full supervision. In this context, weakly supervised, semi-supervised, and sparsely supervised approaches are representative of efficient labeling schemes~\cite{label-efficient-survey}.
The weakly supervised object detection methods~\cite{WS3D, WSS3D, fgr, mixsup} select more lightweight manual annotation, such as clicks, instead of bounding box annotations.
The semi-supervised methods~\cite{SESS,3DIoUMatch,tea-semi,detmatch,hssda, A-Teacher, PatchTeacher} retain annotations for only a subset of frames to train an initial detector, and then employ a teacher-student network to continuously enhance the detection performance of the student model. Existing sparsely supervised methods~\cite{ss3d, CoIn, hinted} further reduce the number of labeled instances, retaining only one labeled instance per annotated frame.
The success of sparsely supervised setting demonstrates that sparse supervision within a scene could support the initialization of detectors.
However, in sparse supervision, the instances labeled per frame are random, endowing the labeled instances with diversity. 
In contrast, the transmitters of shared information among multiple agents are constant, leading to a lack of diversity in the information and thus making the initial detector prone to overfitting to the agents themselves (Fig.~\ref{fig:view}(a)). This motivates us to design an effective 3D object detection framework based on shared information.
\section{Method}
\noindent\textbf{Problem definition.} We start by defining the task of training a detector only from \textit{unlabeled} collaborative driving data, using multi-agent vehicles equipped with synchronized sensors (in particular, LiDAR which provides 3D point clouds, and GPS/INS which provides accurate estimates of vehicle position and orientation).
Such a data collection scheme is practical and annotator-free, making it ideal for cooperative systems where multiple vehicles function together in the same area to gather data collaboratively. Our method automates the annotation of this data, simplifying the workflow and increasing the system's efficiency.

\noindent\textbf{Overview.} 
The pipeline of our \texttt{DOtA} framework is illustrated in Fig.~\ref{fig:framework}. (1) Preliminary Label Generation; (2) Multi-scale Bounding-box Encoding (\textbf{MBE}) for label filtering; (3) Label-Internal Contrastive Learning (\textbf{LICL}). We detail the designs as follows.

\subsection{Preliminary Label Generation}


\begin{table}[t]
	\centering
 \resizebox{0.4\textwidth}{!}{
	\begin{tabular}{c|cc|c|cc}
\toprule
    $\delta$  & \textit{Recall} & \textit{Precision} & $\delta$  & \textit{Recall} & \textit{Precision} \\ \midrule
0.10  &    14.45    &     \textbf{91.93}      & 0.04 &    20.94    &     59.70      \\
0.08  &    15.61    &      89.69     & 0.02 &    29.59    &  23.03         \\
0.06 &   17.54     &   82.34        & 0.01 &     \textbf{47.59 }  &   10.22        \\ \bottomrule
\end{tabular}
    }
    \caption{\textbf{The potential of initial detector.} Utilizing a lower confidence threshold, \textit{e.g.}, $\delta = 0.01$, could result in a label set with a higher recall rate. However, due to the increase in false positive predictions, the precision of the labels declines markedly. Noise-ridden labels hinder the training of effective detectors, underscoring the motivation behind the development of \texttt{DOtA}.}
    \label{tab:motivation_1}
\end{table}

The key idea of \texttt{DOtA} is to utilize the internal shared information of multiple agents and the generalization ability of neural network to obtain a high-recall and high-precision pseudo-label set. To validate the feasibility of this perspective, we conduct a preliminary analysis. 
Specifically, we utilize unlabeled shared information to describe the bounding boxes of agents, and then follow the AttFuse~\cite{opv2v} to train the initial detector with these boxes. Finally, we analyze the inference results of the detector.

Tab.~\ref{tab:motivation_1} shows the recall and precision of the initial labels inferred by initial detector with different confidence score thresholds $\delta$ on the OPV2V~\cite{opv2v} dataset. It is clear to see that the detector is capable of inferring lots of objects in the scene, \textit{e.g.}, achieving a recall rate of $47.95\%$ at a confidence threshold of $0.01$. This results show the potential of the initial-detector only trained with shared information. 
However, during training, due to the absence of positional and shape information of other traffic participants, the detector mistakenly treats other traffic participants as background, leading to an over-fit on agent information. Meanwhile, with the reduction of $\delta$, there is a surge in false positive labels—a significant decrease in precision ( more detail in Fig.~\ref{fig:view}). This motivates us to design
\texttt{DOtA}, a novel framework that effectively expands more valuable information from the shared data, thereby supporting the training of a well-performed detector.

\subsection{Multi-scale Bounding-box Encoding for Label Filtering}
As noted in Tab.~\ref{tab:motivation_1},
in order to maintain a high recall rate, we need to use a lower confidence threshold to preserve the initial labels. 
However, among the retained labels, there is a large number of false positives. To tackle this issue, we introduce the Multi-scale Bounding-box Encoding (\textbf{MBE}) module.
The key idea is to leverage multi-scale and multi-view observation to assess whether the instance points included in the labels adhere to the objective laws of the physical world, thereby retaining high-quality labels.

\noindent\textbf{Bounding-Box Multi-scale Scaling.} Existing unsupervised label filtering methods~\cite{cpd, rffm} typically consider only the point cloud distribution within the bounding box to evaluate the generated labels. However, the contextual information of the labels is equally crucial.
Therefore, for each 3D bounding-box label, $b_{i} = (x_{i},y_{i},z_{i},h_{i},w_{i},l_{i},\theta _{i})$, we slightly scale on it to create new 3D boxes $\left \{ b_{i}^{e\pm} = (x_{i},y_{i},z_{i},h_{i},w_{i}(1 \pm \eta_{e}),l_{i}(1 \pm \eta_{e}),\theta _{i}) \right \}$ to encode the additional information from its context, where $\eta_{e}$ is a constant value for enlarging and reducing the size of box. 
Subsequently, we use the following three context encoding strategies for multi-scale bounding boxes.

\textit{Strategy-1. Collision Probability Encoding.} 
Inspired by \cite{SECOND}, we assume normal traffic participants will not collide with others, meaning there are no points nearby. Therefore, for the enlarged box $b_{i}^{e+}$, we assess the collision probability of the label $b_{i}$ by examining the ratio of point clouds added within the expanded region. A higher ratio indicates a higher probability of collision, suggesting the label is a false positive. For the view of each agent $X^{v}$,  the increase ratio of $b_{i}^{e+}$ is 
\begin{equation}
    r_{i}^{v} = (\left | P_{i}^{e+} \right | - \left | P_{i} \right |)/ \left | P_{i} \right |
\end{equation}
where $\left | P_{i}^{e+} \right |$ , $\left | P_{i} \right |$ are the numbers of points in $b_{i}^{e+}$ and $b_{i}$.

\textit{Strategy-2. Boundary Alignment Encoding.} Due to the nature of LiDAR sensing, the edges of the inner-points should align with the boundaries of the label. To determine the edges, we employ the Qhull~\cite{qhull} algorithm to obtain the convex hull on the interior points, using it as a reference for the edge points. Therefore, we slightly shrink the label, and we assess the degree of alignment by calculating the ratio that the convex hull falls within the reduced area. 
A lower ratio indicates a worse alignment of the label, suggesting the label is low-quality. For the view of each agent $X^{v}$,  the occupancy ratio of $b_{i}^{e-}$ is 
\begin{equation}
    o_{i}^{v} = (\left | Q_{i} \right | - \left | Q_{i}^{e-} \right |)/ \left | Q_{i} \right |
\end{equation}
where $\left | Q_{i}^{e-} \right |$ and $\left | Q_{i} \right |$ are the numbers of convex hull in $b_{i}^{e-}$ and $b_{i}$.

\textit{Strategy-3. Information Confidence Encoding.} For LiDAR scanning, the point cloud is dense near and sparse far away, hence the agents closer to the label provide more information. Therefore, we describe the confidence of information based on the Euclidean distance between the label $b_{i}$ and the agent $X_{v}$:
\begin{equation}
    d_{i}^{v} = 1 / [(x_{v}- x_{i})^{2} + (y_{v}-y_{i})^{2}]
\end{equation}
where $(x_{v},y_{v})$ is the 2D center of the agent $X^{V}$. The larger the encoded $d_{i}^{v}$, the higher the confidence.

We perform MBE from the view of each agent, resulting in: $\left \{ [r_{i}^{v}, o_{i}^{v}, d_{i}^{v}] \mid v = 1,...,V \right \}$, where $V$ is the number of agents.
And then, by combining the encoded information relayed by all agents, we design a label discriminator: 
\begin{equation}
  \mathcal{D} = \left\{\begin{matrix}
  condition_{1}: \quad \left( \sum_{v=1}^{V} r_{i}^{v} \times \frac{d_{i}^{v}}{\sum_{v'=1}^{V} d_{i}^{v'}} \right) < \varphi_{r} \\  
  condition_{2}: \quad \left( \sum_{v=1}^{V} o_{i}^{v} \times \frac{d_{i}^{v}}{\sum_{v'=1}^{V} d_{i}^{v'}} \right) > \varphi_{o}
  \end{matrix}\right.
\end{equation}
where $\varphi_{r}$ and $\varphi_{o}$ indicate the collision tolerance and alignment tolerance, respectively. A label is assigned as high-quality if it simultaneously meets the two conditions of the discriminator; otherwise, it is considered a low-quality label.
Only high-quality labels will be directly used as supervisory signals to participate in the training of the classification and regression modules of the detection model.

\subsection{Label-Internal Contrastive Learning}
By combining information from multiple agents, we obtain distinguishable labels. However, low-quality labels are also the outcomes of the detector, mapping the erroneous feature learning from the previous round. To correct this misguided feature learning and to emphasize the correct ones from the previous round, we propose a Label-Internal Contrastive Learning (LICL) module. The key idea is to introduce contrastive learning~\cite{moco}, using historical predictions as cues to enhance current feature learning capabilities.

We firstly extract local feature for each promising label $\left \{ b_{i} = (x_{i},y_{i},z_{i},h_{i},w_{i},l_{i},\theta _{i}) \right \}$ by leveraging the correlation between the resolution of the feature map ($W \times H$) and the range of the original scene point cloud ($\left [x_{min} : x_{max}, y_{min} : y_{max} \right ] $).
\begin{equation}
    index_{i} = \left ( \lfloor \frac{x_{i}-x_{min}}{x_{max}-x_{min}}\rfloor   \times  W , \lfloor \frac{y_{i}-y_{min}}{y_{max}-y_{min}}\rfloor   \times  H  \right ). 
\end{equation}
where $index_{i}$ represents the position index of the region corresponding to $b_{i}$ in the feature map. Based on the MBE module, according to the quality of the label $b_{i}$, we further identify the local feature $F \left ( index_{i} \right )$ into positive feature $ f_{pos} $ and negative feature $ f_{neg} $ . 


For all high-quality labels $\left\{ b_{pos}^{m}, m=1,...,M \right\} $ and low-quality labels $\left\{b_{neg}^{n}, n=1,...,N \right\} $, we can generate corresponding sets of positive features $\left\{ f_{pos}^{m}, m=1,...,M \right\} $ and negative features $\left\{ f_{neg}^{n}, n=1,...,N \right\} $. Following existing efforts~\cite{CoIn, pointcontrast}, we apply the InfoNCE~\cite{infonce} loss to pull closer between positive features and to push positive and negative features further apart within the feature space.
\begin{equation}
\mathcal{L}_{LICL} = \frac{-1}{M} \sum_{m = 0}^{M}  log\frac{exp\left ( \sum_{i = 0}^{M} \left ( f_{pos}^{m} \cdot f_{pos}^{i} \right )/ \tau  \right ) }{\sum_{n=0}^{N} exp \left ( \sum_{i = 0}^{M} \left ( f_{pos}^{i} \cdot f_{neg}^{n} \right ) /\tau  \right ) } .
\label{eq2}
\end{equation}
where $\tau$ is a temperature scaling parameter \cite{ufl}. The LICL module utilizes high-quality and low-quality labels as prompts to encourage correct feature learning from the previous round and suppress erroneous feature learning. It is worth noting that this module is only involved in the model training to enhance the feature learning of the backbone and does not participate in the inference phase.


\subsection{Training Losses}
Following previous collaborative perception methods~\cite{how2comm, where2comm}, we utilize the smooth absolute error loss for regression, denoted as  $\mathcal{L}_{reg}$, and the focal loss~\cite{FocalLoss} for classification, denoted as $\mathcal{L}_{cls}$. In total, our proposed \texttt{DOtA} framework defines the comprehensive objective function as follows:
\begin{equation}
\mathcal{L}_{total} = \alpha \mathcal{L}_{reg} + \beta \mathcal{L}_{cls} + \gamma \mathcal{L}_{LICL}.
\label{eq3}
\end{equation}
where $\alpha$ and $\beta$ are empirically set to $1$ according to~\cite{opv2v}, hyper-parameter $\gamma$ balances the task of detection and feature learning. We conduct the ablation study for $\gamma$ in section~\ref{ablation}.

\section{Experiments}

\begin{table*}[htbp]
	\centering
     \resizebox{0.9\textwidth}{!}{
     \begin{tabular}{l|c|c|c|cccccc}
\toprule
\multirow{2}{*}{Methods} & \multirow{2}{*}{Reference} & \multirow{2}{*}{\begin{tabular}[c]{@{}c@{}}Common \\ sense\end{tabular}} & \multirow{2}{*}{\begin{tabular}[c]{@{}c@{}}Internal\\ shared-info\end{tabular}} & \multicolumn{3}{c}{V2V4real~\cite{v2v4real}} & \multicolumn{3}{c}{OPV2V~\cite{opv2v}} \\
                         &                            &                                                                          &                                                                                 & AP@0.3       & AP@0.5 &  AP@0.7    & AP@0.3        & AP@0.5  &  AP@0.7     \\ \midrule \hline
\multirow{2}{*}{DBSCAN~\cite{dbscan}}  & \multirow{2}{*}{KDD 1996}  &                                        \checkmark                                   &                                                                  & 9.26          & 4.39   &  0.43               & 29.49         & 22.24  &    9.38       \\
                         &                            &           \checkmark                                                                &                                                        \checkmark                     & 14.59          & 7.57  &0.88     & 32.31         & 24.43   &   11.63          \\ \hline
\multirow{2}{*}{MODEST~\cite{MODEST}}  & \multirow{2}{*}{CVPR 2022} &                                      \checkmark                                     &                                                                          & 24.13          & 12.55     &1.83       & 46.61         & 34.27   &   14.96       \\
                         &                            &                                                      \checkmark                     &                                                             \checkmark             & 33.44          & 23.32   &3.60            & 50.83         & 43.43    &  20.58     \\ \hline
\multirow{2}{*}{OYSTER~\cite{OYSTER}}  & \multirow{2}{*}{CVPR 2023} &                                      \checkmark                                     &                                                                 & 16.61          & 7.47    &0.36                & 51.92         & 47.27   &  20.18         \\
                         &                            &                                   \checkmark                                       &                                                   \checkmark                        & 37.50          & 23.52  &4.84       & 56.58         & 49.01  &   24.34           \\ \hline
\multirow{2}{*}{CPD~\cite{cpd}}     & \multirow{2}{*}{CVPR 2024} &                                                           \checkmark                &                                                                     & 35.55          & 27.06    &3.76            & 54.83         & 48.34       & 23.56      \\
                         &                            &                                            \checkmark                               &                                                           \checkmark            & 40.67          & 30.27    &4.73           & 59.17         & 50.49   &  \textbf{27.72}         \\ 
                         \hline
                         \midrule 
DOtA  (Ours )                   &  -                          &                                                                          &                      \checkmark                                                      & \textbf{54.60}          & \textbf{48.84}  &   \textbf{22.41}      & \textbf{66.14}         & \textbf{52.37} &    24.57        \\ \bottomrule
\end{tabular}
}
    \caption{Comparison with unsupervised methods on OPV2V dataset~\cite{opv2v} and V2V4Real dataset~\cite{v2v4real}. All methods are based on AttFuse~\cite{opv2v}. We report the results of Average Precision (AP) under Intersectionover-Union (IoU) 0.3, 0.5 and 0.7. The best performance are highlighted in \textbf{bold}.}
    \label{tab:com_unsupervised}
\end{table*}

\subsection{Datasets and Evaluation Metrics} 
We conduct experiments with two different datasets: V2V4Real dataset~\cite{v2v4real} and OPV2V dataset~\cite{opv2v}. To the best of our knowledge, these two are representative collaborative perception datasets, one of which is a real dataset collected from large-scale real-world scenarios, and the other is a virtual dataset collected based on simulation emulators~\cite{Xu_Guo_Han_Xia_Xiang_Ma_2021, Dosovitskiy_Ros_Codevilla_López_Koltun_2017}. We validate the effectiveness of \texttt{DOtA} on both types of datasets. To ensure a fair comparison, we utilized the official evaluation metric:
the Average Precision (AP) under Intersection over Union (IoU) $0.3, 0.5$ and $0.7$.


\subsection{Implementation Details}
\label{detail}

\noindent \textbf{Training Details.} 
In multi-agent collaborative perception systems, each agent transmits its pose data. Concurrently, details of the collaborating vehicles, including their types, are pre-registered~\cite{v2v4real}. 
Consequently, we could easily obtain the internal shared-info of each agent vehicle $X^{v}$ in the scene without manual labeling: $(x_{v}, y_{v}, z_{v},h_{v},w_{v},l_{v},\theta_{v})$.
During the initialization phase of the detector, we directly utilize these internal shared-info from the agents as supervision. In the design of the discriminator $\mathcal{D}$, the collision tolerance parameter $\varphi_{r} $ is set to 0.1, and the alignment tolerance parameter $\varphi_{o} $ is set to 0.7. Additionally, scaling factor $\eta_{e}=[0.5, 0.2]$. 

\noindent \textbf{Baseline Details.} We are the first to develop a method for training collaborative detectors only with internal
shared-info, and there are no previously published baselines for comparison. To validate the effectiveness of \texttt{DOtA}, we chose to compare it with works~\cite{MODEST, OYSTER, cpd} that are also without manual labels.
To ensure a fair comparison, we adopt widely used PointPillars~\citep{PointPillars} as basic detector. Furthermore, we also pass internal shared-info to previous unsupervised methods, enhancing their detection capabilities. 

\begin{table*}[htbp]
	\centering
     \resizebox{0.8\textwidth}{!}{
     \begin{tabular}{l|c|c|cccc}
\toprule 
\multirow{2}{*}{Methods} & \multirow{2}{*}{\begin{tabular}[c]{@{}c@{}}Manual \\ Labels\end{tabular}} & \multicolumn{1}{l|}{\multirow{2}{*}{\begin{tabular}[c]{@{}c@{}}Internal\\ shared-info\end{tabular}}} & \multicolumn{2}{c}{V2V4real~\cite{v2v4real}} & \multicolumn{2}{c}{OPV2V~\cite{opv2v}} \\
                         &                                                                           & \multicolumn{1}{l|}{}                                                                                & AP@0.3      & AP@0.5      & AP@0.3       & AP@0.5      \\ \midrule \hline
AttFuse~\cite{opv2v}                  &                                     \checkmark                                      &                                                                                         & 71.35          & 63.15             & 85.50         & 83.21                  \\
AttFuse~\cite{opv2v}               & \multirow{2}{*}{}                                                         & \checkmark                                                                             & 11.15          & 10.62        & 11.75         & 11.55          
\\
AttFuse~\cite{opv2v} + self-training~\cite{ss3d}               & \multirow{2}{*}{}                                                         & \checkmark                                                                    &    20.38       & 16.47                &   30.02       &    18.18     
\\
DOtA based AttFuse       &                                                                           &                                                                  \checkmark                           & \textbf{54.60}          & \textbf{48.84}           & \textbf{66.14}         & \textbf{52.37}                 \\ \hline
V2X-ViT~\cite{v2x-vit}                   &                                        \checkmark                                    &                                                                                         & 72.10          & 62.87             & 85.99         & 84.13                  \\
V2X-ViT~\cite{v2x-vit}                  & \multirow{2}{*}{}                                                         & \checkmark                                                                     & 12.26          & 11.98                & 7.81         & 7.75           
\\
V2X-ViT~\cite{v2x-vit}+self-training~\cite{ss3d}                  & \multirow{2}{*}{}                                                         & \checkmark                                                                           &  18.43         & 14.30           &       33.05   &    19.77     
\\
DOtA based V2X-ViT      &                                                                           &                                                                   \checkmark                           & \textbf{61.90}          & \textbf{54.45}          & \textbf{64.33}         & \textbf{46.06}                 \\ \hline
Where2Comm~\cite{where2comm}               &                                       \checkmark                                     &                                                                                       & 74.02          & 66.80                & 87.03         & 84.40                 \\
Where2Comm~\cite{where2comm}               & \multirow{2}{*}{}                                                         & \checkmark                                                                         & 11.82          & 6.89              & 7.61         & 7.31                \\
Where2Comm~\cite{where2comm}+self-training~\cite{ss3d}               & \multirow{2}{*}{}                                                         & \checkmark                                                                          &  20.04         &   13.53           &    31.88      &    20.14            \\
DOtA based Where2Comm    &                                                                           &                                                                  \checkmark                       & \textbf{51.55}          & \textbf{43.96}              & \textbf{66.62}         & \textbf{53.23}                  \\ \bottomrule
\end{tabular}
}
    \caption{ Verification on different collaborative methods with full manual labels and internal shared-info on OPV2V dataset~\cite{opv2v} and V2V4Real dataset~\cite{v2v4real}. We report the results of Average Precision (AP) under Intersectionover-Union (IoU) 0.3 and 0.5. The best performance only with internal shared-info are highlighted in \textbf{bold}.
    }
    \label{tab:com_fusion}
\end{table*}

\subsection{Main Results}
\noindent \textbf{Comparison with unsupervised methods.} We compare the proposed \texttt{DOtA} with the most advanced unsupervised methods~\cite{dbscan,MODEST,OYSTER,cpd}. 
For a fair comparison, all methods adopt PointPillars~\cite{PointPillars} as the backbone and AttFuse~\cite{opv2v} as the fusion strategy. Tab.~\ref{tab:com_unsupervised} shows a performance comparison of two public datasets. 

For the V2V4Real dataset~\cite{v2v4real}, since the data is collected from real-world scenarios, there is noise present in the data due to communication delays and localization errors. This noise prevents the early-fused point clouds from aligning well in 3D space. Therefore, to ensure label quality, we employ a late-fusion strategy to generate labels for the baseline unsupervised methods~\cite{dbscan, MODEST,OYSTER,cpd}. 
The experimental results on V2V4Real are shown on the left side of Tab.~\ref{tab:com_unsupervised}. Relying solely on internally shared information, our DOtA surpasses previous methods at all IoU thresholds. Additionally, traditional unsupervised schemes are susceptible to noise in real-world data, leading to a significant reduction in detector performance at the 0.7 IoU threshold. In contrast, our method does not rely on cluster-based bounding box fitting strategies. Still, it leverages the generalized performance of neural networks to expand labels, thus exhibiting greater robustness against these noises.

For the OPV2V dataset~\cite{opv2v}, to capitalize on the strengths of collaborative perception datasets and to achieve optimal performance with traditional unsupervised methods~\cite{MODEST, dbscan, OYSTER, cpd}, we use early-fusion point clouds as the input for these methods. Notably, to reduce computational load, \texttt{DOtA} adopts an intermediate-fusion approach by using individual agent point clouds as input separately, ultimately only merging the encoded information.
The experimental results on OPV2V are shown in the right side of Tab.~\ref{tab:com_unsupervised}. 
Our \texttt{DOtA} surpasses the performance of all previous unsupervised methods at IoU thresholds of 0.3 and 0.5, without relying on common-sense information. Compared to the previously best-performing method CPD, our approach lags slightly under the IoU threshold of 0.7. This is due to the fact that OPV2V is a simulation dataset with less internal noise, which is more conducive to the performance of traditional unsupervised methods.

\noindent \textbf{Validation with different collaborative methods.}
We select three classic collaborative perception object detection methods, AttFuse~\cite{opv2v}, V2X-VIT~\cite{v2x-vit}, and Where2Comm~\cite{where2comm}, as baselines. Then, we train the baseline detection models with the internal shared-info and the full manual labeling, separately. The results are reported in Tab.~\ref{tab:com_fusion}. Due to the limited amount of information in the internally shared-info, the performance of the detector directly trained in this manner is relatively poor. We combine a self-training scheme, commonly used in unsupervised strategies~\cite{MODEST,cpd}, into our baseline method, which results in a certain improvement in performance. However, such self-training methods struggle to strictly control label quality, hence the performance of the detector still lags significantly behind that of a fully supervised detector. By adding our MBE and LICL to the three baseline methods, our \texttt{DOtA} pipeline achieved an average performance improvement of $32.61\%$ and $35.35\%$ on two datasets, respectively. This further narrows the gap with fully supervised methods.

\begin{table}[htbp]
\resizebox{\columnwidth}{!}{
\begin{tabular}{c|cc|cc|cc}
\toprule
\multirow{2}{*}{Method} & \multicolumn{2}{c|}{Car} & \multicolumn{2}{c|}{Ped.} & \multicolumn{2}{c}{Truck} \\
                        & Ap@0.3        & Ap@0.5       & Ap@0.3         & Ap@0.5        & Ap@0.3      & Ap@0.5      \\ \midrule
CPD                     &       27.56        &       20.67       &    11.37            &   7.52             &   13.32          &   10.88           \\
LISO                    &     26.89          &       20.38       &    9.03            &    5.42          &     14.33        &     11.62       \\
DOtA                    &         34.27      &     26.96         &        14.33        &      10.25         &  17.67           &    14.56         \\ \bottomrule
\end{tabular}

}
\caption{Comparison of unsupervised methods on V2X-Real~\cite{V2x-real}.}
    \label{tab:multi-class}
\end{table}

\noindent \textbf{Results  of  Multi-Class Detection.} As shown in Tab~\ref{tab:multi-class}, we compare our DOtA with unsupervised methods on V2X-Real dataset~\cite{V2x-real}. V2X-Real is a real-world multi-class 3D detection cooperative perception dataset. Following mainstream strategies~\cite{cpd}, we use size priors to classify class-agnostic labels. Benefiting from the effective utilization of collaborative data, DOtA achieves the optimal performance.

\subsection{Label Quality Analysis}
\begin{figure}[t]
  \centering
   \includegraphics[width=1\linewidth]{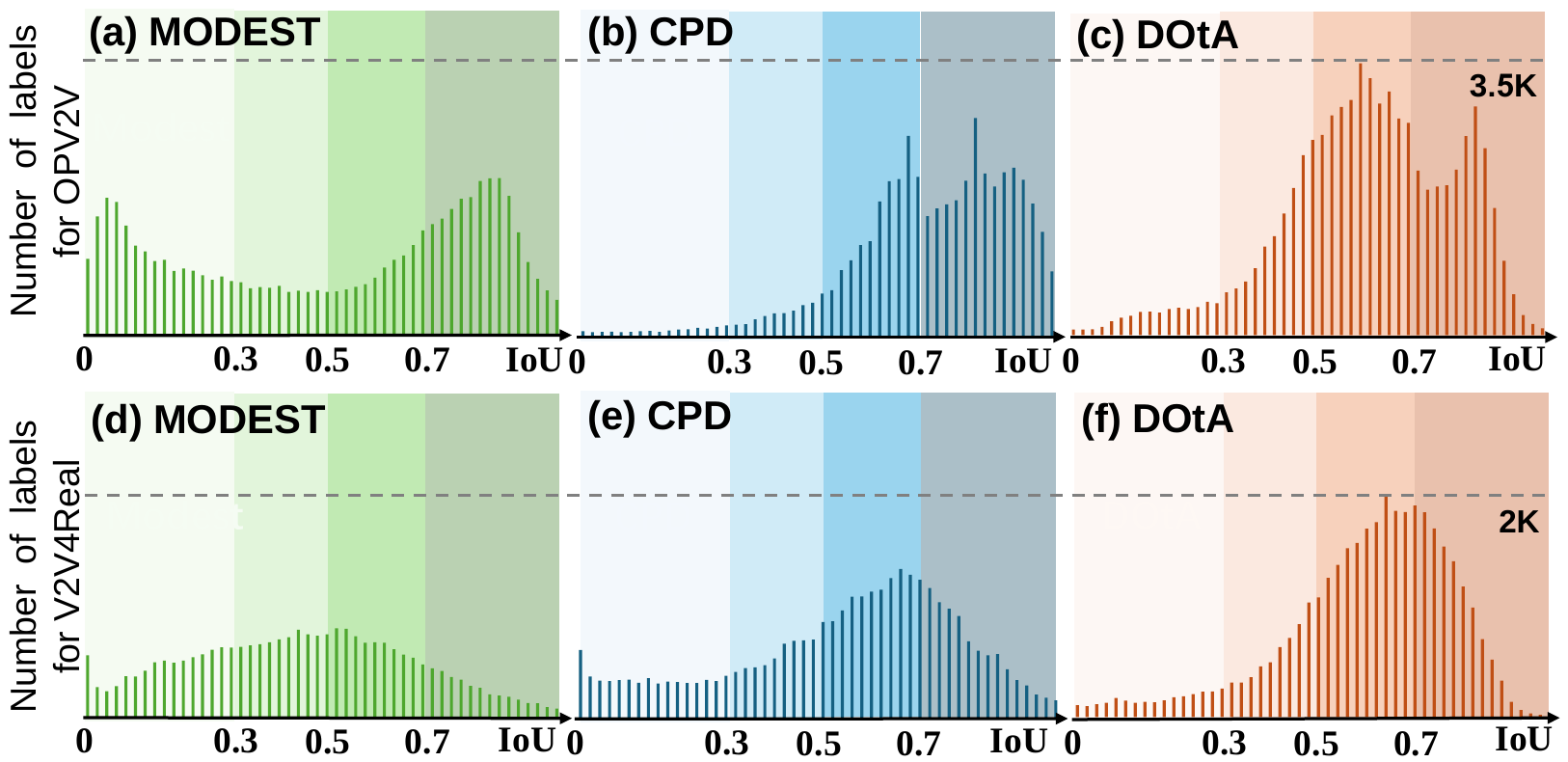}
   \caption{The IoU distribution between pseudo-labels and ground truth is presented, where figures (a), (b), and (c) correspond to OPV2V, and figures (d), (e), and (f) correspond to V2V4Real.
   }
   \label{fig:label_quality}
\end{figure}

To verify the labels generated by \texttt{DOtA}, we use the ground-truth as a reference to analyze recall and precision on OPV2V~\cite{opv2v} and V2V4Real~\cite{v2v4real} datasets.
As shown in Tab.~\ref{tab:label_quality}, the label quality of \texttt{DOtA} significantly surpasses that of previous unsupervised methods.
Because the initial labels of \texttt{DOtA} are generated based on the generalization of neural networks, they have stronger noise resistance than traditional unsupervised methods based on clustering fitting. Therefore, on real datasets, the label quality of \texttt{DOtA} is more robust.
To understand the sources of this improvement, we examined the IoU between the pseudo-labels and ground truth, and compared the IoU distributions in Fig.~\ref{fig:label_quality}. From the figure, it can be observed that \texttt{DOtA} maintains a similar IoU distribution on both synthetic and real-world datasets.

\begin{table}[htbp]
	\centering
\resizebox{0.48\textwidth}{!}{
   \begin{tabular}{l|cccc}
\toprule
\multirow{3}{*}{Method} & \multicolumn{2}{c}{V2V4Real~\cite{v2v4real}}   & \multicolumn{2}{c}{OPV2V~\cite{opv2v}} \\ \cline{2-5} 
                        & Recall       & Precision    & Recall        & Precision    \\
                        & IoU @0.3/0.5 & IoU @0.3/0.5 & IoU @0.3/0.5  & IoU @0.3/0.5 \\ \midrule
DBSCAN~\cite{dbscan}        & 32.75/19.59   & 4.20/2.51          & 47.16/39.54  & 14.11/11.83    \\
MODEST~\cite{MODEST}          & 31.33/19.00   & 13.43/8.14        & 37.06/31.01  & 58.52/48.96    \\
OYSTER~\cite{OYSTER}            & 41.05/29.32   & 22.04/15.74      & 48.35/43.72  & 60.69/54.89    \\
CPD~\cite{cpd}          & 40.52/32.15   & 20.75/16.46           & 48.30/45.80  & 59.31/56.23    \\
DOtA (Ours)           & \textbf{52.31/43.91}   & \textbf{71.97/60.42}         & \textbf{62.60/51.87}  & \textbf{79.34/65.74}    \\ \bottomrule
\end{tabular}
    }
    \caption{The comparison of label quality on OPV2V~\cite{opv2v} and V2V4Real~\cite{v2v4real} datasets.}
    \label{tab:label_quality}
\end{table}

\begin{figure*}[t]
  \centering
   \includegraphics[width=1\linewidth]{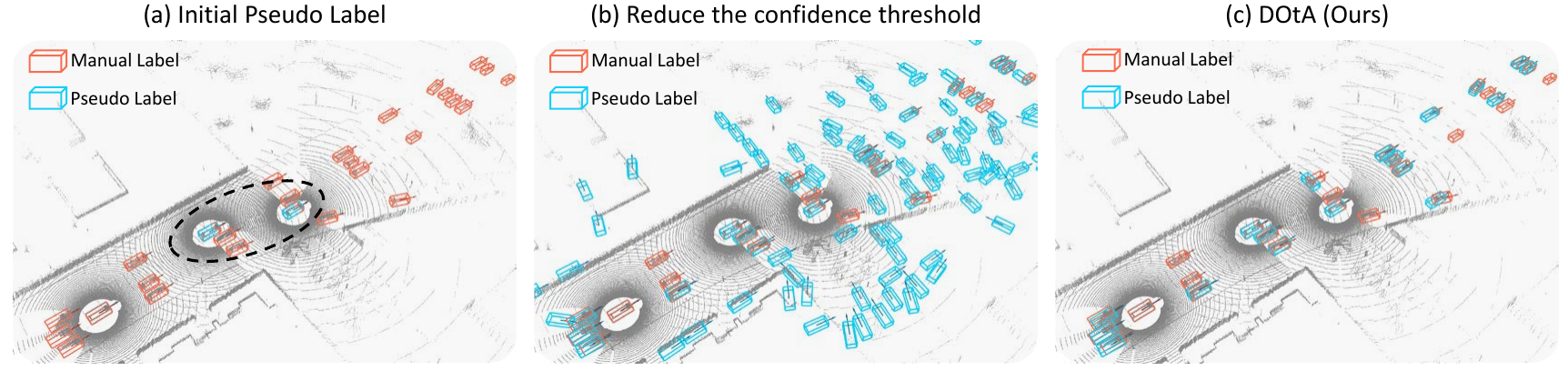}
   \caption{ Visualization of the label filtering process of DOtA on the OPV2V $train$ split.}
   \label{fig:view}
\end{figure*}

\subsection{Ablation Study}
\label{ablation}

\noindent \textbf{Effect of multi-scale bounding-box encoding.} The first through seventh rows of Tab.~\ref{tab:aba_three_module} report the effects of Multi-scale Bounding-box Encoding (MBE). MBE encompasses three distinct strategies: Collision Probability Encoding (CPE), Boundary Alignment Encoding (BAE), and Information Confidence Encoding (ICE). We perform ablation studies on them sequentially. In the $1^{st}$ row of Tab.~\ref{tab:aba_three_module}, we report the performance of the AttFuse~\cite{opv2v} with self-training as our baseline. In the $2^{nd}$ row, we add CPE to filter labels, resulting in a significant performance improvement. This demonstrates that the CPE module eliminates a large number of false positives and enhances the accuracy of the labels. In the $3^{rd}$ row, we use only BAE to filter labels, which also brings about a certain performance improvement. By combining CPE and BAE, our DOtA further improves the performance. Taking into account the varying contributions of each agent, we proposed the ICE. The results in rows $5^{th}, 6^{th}$, and $7^{th}$ demonstrate that the ICE module can assign an appropriate degree of contribution to each agent, ultimately producing the highest quality labels.

\begin{table}[htbp]
	\centering
\resizebox{0.4\textwidth}{!}{
    \begin{tabular}{c|ccc|c|cc}
\toprule
\multirow{2}{*}{ID} & \multicolumn{3}{c|}{MBE}   & \multirow{2}{*}{LICL} & \multicolumn{2}{c}{OPV2V~\cite{opv2v}} \\  
& CPE & BAE & ICE &                      & AP@0.3      & AP@0.5      \\ \midrule
        
1  &      &       &          &                      &     30.02    &    18.18     \\
2  &   \checkmark     &       &          &                      & 55.62        & 40.01        \\
3  &        &   \checkmark    &          &                      & 43.28        & 24.79        \\
4  &    \checkmark    &   \checkmark    &          &                      & 58.75        & 43.27       \\
5  &     \checkmark   &       &     \checkmark     &                      & 57.63        & 40.98        \\
6  &        &     \checkmark  &    \checkmark      &                      & 48.56        & 26.34        \\
7  &    \checkmark    &  \checkmark     &     \checkmark     &                      & 60.96        & 45.77        \\
8  &      \checkmark  &    \checkmark   &    \checkmark      &  \checkmark                    & 66.14        & 52.37        \\ \bottomrule
\end{tabular}
    }
    \caption{Effects of the different components on our designed DOtA network. }
    \label{tab:aba_three_module}
\end{table}

\noindent \textbf{Effect of label-internal contrastive learning.}
As shown in the $7^{th}$ and $8^{th}$ rows from Tab.~\ref{tab:aba_three_module}, whether the use of label-internal contrastive learning (LICL) module has a certain impact on the performance of the detector. By combining LICL with MBE to train the detector, our DOtA boosts the performance of $AP@0.3$ and $AP@0.5$ by about $5.18$ and $6.60$ percentage points, respectively, as shown in the $7^{th}$ and $8^{th}$ rows. This verifies the effectiveness of the jointly label inner contrastive learning strategy.


\begin{table}[htbp]
	\centering
        \resizebox{0.45\textwidth}{!}{
	\begin{tabular}{c|ccccccc}
\toprule
$\gamma$ & 0.1  & 0.3  & 0.5  & 0.7  & 1.0  & 1.5  & 2.0  \\ \midrule
AP@0.3 & 0.61 & 0.61 & 0.62 & 0.63 & \textbf{0.66} & 0.63 & 0.63 \\ \bottomrule
\end{tabular}
}
\caption{The choice of weight parameter $\gamma$ of $\mathcal{L}_{LICL}$. }
    \label{tab:hy_select}
\end{table}

\noindent \textbf{Choice of the hyper-parameter $\gamma$.} The setting of $\gamma$ determines the extent of the influence of the LICL module throughout the entire training process. In this section,  we investigate the selection of $\gamma$ of the optimal loss $\mathcal{L}_{LICL}$. From the results in Tab.~\ref{tab:hy_select}, it is observed that the LICL module achieves maximum performance when $\gamma$ is set to $1.0$.

\noindent \textbf{Label-filtering Visualization}
\label{viii}
To provide a more detailed depiction of the label filtering process in \texttt{DOtA} and its underlying motivation, we offer a comprehensive visualization. Firstly, we set a confidence threshold of $0.1$ to filter the output of the inference of initial detector on the $train$ split. As shown in Fig.~\ref{fig:view} (a), the inference results below this threshold only cover the agent, indicating that the model is overfitting to the agent. And then, we set a lower confidence threshold of $0.01$, the result is shown in Fig.~\ref{fig:view} (b). 
At this point, the inferred labels can cover more ground-true bounding boxes, but a large number of false positives have emerged. To eliminate false positives and retain more high-quality labels, we design DOtA, and the final labels are shown in Fig.~\ref{fig:view} (c).

\section{Conclusion}
In this paper, to overcome the limitations of traditional unsupervised 3D object detection methods that are constrained by a single-view observation,
we proposed a novel detect objects from multi-agent LiDAR scans method, \texttt{DOtA}, without manual labels.  
Instead of requiring cluster-based label-fitting methods, our method utilizes the internal shared-info to initialize an object detector. In this way, our \texttt{DOtA} method does not require the meticulous design of hyper-parameters based on common-sense information to generate initial labels, as traditional unsupervised methods do, thereby enhancing the robustness. 
To improve the performance of detector, we designed a multi-scale bounding box encoding module to filter the initial labels, ultimately retaining those of high quality to train detector. And then, we propose a label-internal contrastive learning module that guides the detector to better accomplish feature learning. 
Extensive experiments have verified the effectiveness of our design.

\noindent \textbf {Acknowledgment.} This work was supported by the National Natural Science Foundation of China (No.62171393), and the Fundamental Research Funds for the Central Universities (No.20720220064).

{
    \small
    \bibliographystyle{ieee_fullname}
    \bibliography{main}
}


\end{document}


\maketitle


\begin{figure*}[h]
  \centering
   \includegraphics[width=1\linewidth]{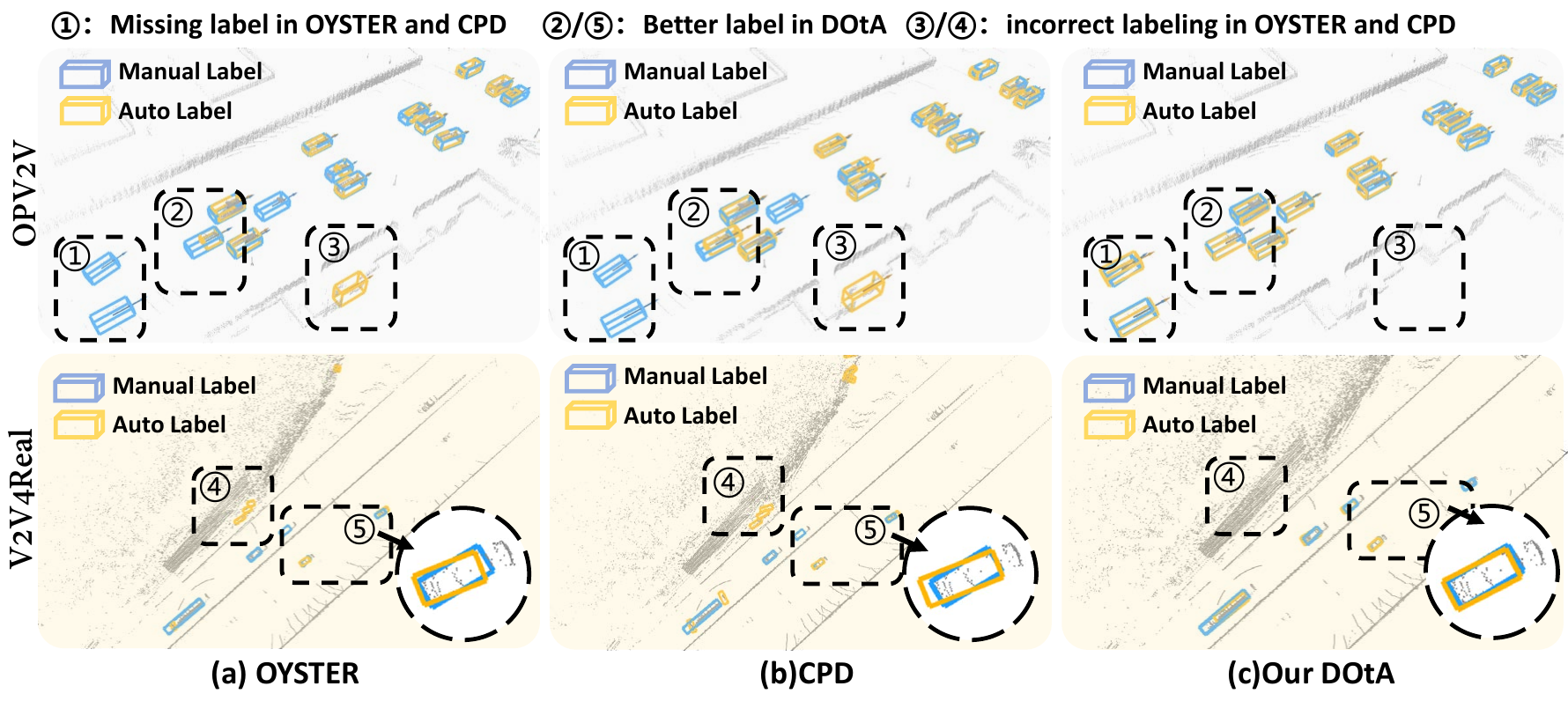}
   \caption{ Visualization comparison of different labels on OPV2V $train$ split.}
   \label{fig:view_compare}
\end{figure*}

\section{Label visualization comparison of different methods on two datasets.}
As shown in the Fig.~\ref{fig:view_compare}, we present a visual comparison of the automatic labeling results for different unsupervised methods on both the simulated dataset OPV2V~\cite{opv2v} (top) and the real-world dataset V2V4Real~\cite{v2v4real} (bottom). For ease of observation, we have removed the ground point cloud from the original point cloud scene. In the simulated dataset OPV2V, within the areas under multi-agent synchronous observation, both existing unsupervised approaches and our approach can achieve commendable results in automatic labeling. 
However, objects located at the edge of the collaborative observation area are still affected by occlusion and the sparsity of point clouds, leading to incomplete structures in the point cloud descriptions. This phenomenon can degrade the performance of traditional unsupervised clustering-based algorithms. For instance, in area \textcircled{1}, due to the limited range of the point clouds output by clustering, traditional methods may discard the fitted bounding boxes for this part; in areas \textcircled{2} and \textcircled{5}, the incomplete structures resulting from clustering lead to poor bounding box fitting effects. Additionally, traditional fitting-based methods only consider the dimensions of the fitting box for selecting bounding boxes, which can lead to incorrect labels,\textit{e.g.}, in many areas labeled as \textcircled{3} and \textcircled{4}.

\section{Additional Study on Robustness to localization noise.}
\begin{figure*}[h]
  \centering
   \includegraphics[width=0.5\linewidth]{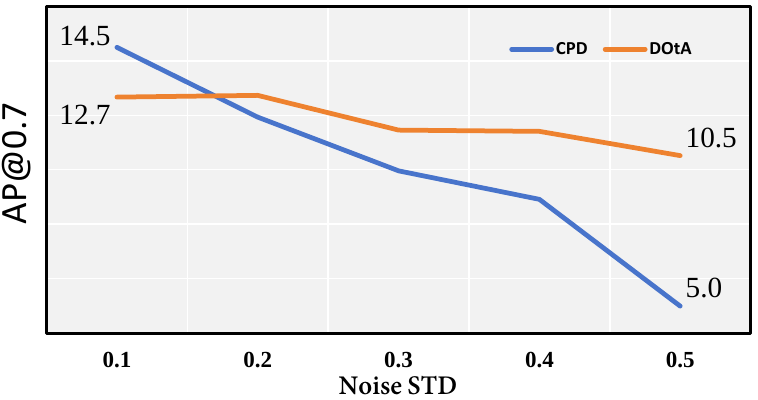}
   \caption{ Additional Study on Robustness to localization noise.}
   \label{fig:noise_ro}
\end{figure*}

Due to localization errors and communication delays, there is noise in real-world multi-agent collaborative observation. To further investigate the impact of this issue on our method, we follow the localization noise setting in Where2comm~\cite{where2comm} (Gaussian noise with a mean of $0m$ and a standard deviation of $0m$-$0.6m$) and conduct experiments on OPV2V dataset to validate the robustness against realistic localization noise. As the results shown in Fig.~\ref{fig:noise_ro}, compared to the traditional unsupervised method CPD~\cite{cpd}, our DOtA demonstrates stronger robustness. To further investigate the sources of robustness in DOtA, we statistically analyze the distribution of Intersection over Union (IOU) between DOtA labels and the ground truth labels, as depicted in the Fig.~\ref{fig:noise}.

\begin{figure*}[h]
  \centering
   \includegraphics[width=0.8\linewidth]{figures/iou_distribution(1).pdf}
   \caption{ Distribution of Intersection over Union (IOU) between DOtA labels and the ground truth labels.}
   \label{fig:noise}
\end{figure*}

\section{The study of collision tolerance parameter $\varphi_{r}$ and alignment parameter $\varphi_{o}$.}
In this section, we study the impact of different collision tolerance parameter and alignment parameter on detector performance. As shown in Tab.~\ref{cta}, we final set $\varphi_{r} = 0.1$ and $\varphi_{o} = 0.7$.
\begin{table}[h]
\centering
\begin{tabular}{cccccccccccccc}
\cline{1-6} \cline{8-14}
$\varphi_{r}$      & 0.01 & 0.05 & 0.10 & 0.15 & 0.20 &  & $\varphi_{o}$      & 0.1 & 0.3 & 0.5 & 0.7 & 0.8 & 0.9 \\ \cline{1-6} \cline{8-14} 
AP@0.5   & 20.66 & 35.27 & \textbf{40.01} & 38.25 & 37.63 &  & AP@0.5 & 40.23 & 40.79 & 41.45 & \textbf{43.27} & 43.08 & 42.77 \\ \cline{1-6} \cline{8-14} 
\end{tabular}
\caption{The study of collision tolerance parameter $\varphi_{r}$ and alignment parameter $\varphi_{o}$.}
\label{cta}
\end{table}

\section{The study of scaling factor $\eta_{e}$.}
In this section, we study the impact of different scaling factor on label recall rates. As shown in Tab.~\ref{scale}, the optimal scaling factors are [0.5, 0.2].

\begin{table}[h]
\centering
\begin{tabular}{cccccccc}
\hline
$\eta_{e}$      & {[}0.4, 0.2{]} & {[}0.4, 0.3{]} & {[}0.5, 0.2{]} & {[}0.5, 0.3{]} & {[}0.6, 0.2{]} & {[}0.6, 0.3{]} & {[}0.6, 0.4{]} \\ \hline
Reacll@0.5 &         51.66       &        51.45        &       \textbf{51.87}         &         51.73       &         51.61        &      51.70          &      50.21          \\ \hline
\end{tabular}
\caption{The study of scaling factor $\eta_{e}$.}
\label{scale}
\end{table}

{
    \small
    \bibliographystyle{ieee_fullname}
    \bibliography{egbib}
}
